\documentclass[10pt,conference]{IEEEtran}
\IEEEoverridecommandlockouts

\usepackage{cite}
\usepackage{amsmath,amssymb,amsfonts}
\usepackage{algorithmic}
\usepackage{booktabs}
\usepackage{graphicx}
\usepackage{textcomp}
\usepackage{xcolor}
\usepackage[normalem]{ulem}
\usepackage{amsmath}
\usepackage{booktabs}
\usepackage{tabularx}
\usepackage{multirow}

\usepackage{tikz}
\usetikzlibrary{positioning,arrows.meta}
\usepackage{forest}

\usepackage[color=green, size=scriptsize]{todonotes} 

\usepackage{hyperref}
\hypersetup{
 colorlinks=true,
 linkcolor=blue,
 filecolor=blue,
 citecolor = blue,      
 urlcolor=blue,
}

\usepackage[numbers]{natbib}
\usepackage{url}

\usepackage[font={small, color={black}}]{caption}
\usepackage{subcaption}

\usepackage{enumitem}
\setlist[itemize]{noitemsep, topsep=0.5pt, parsep=0pt, partopsep=0pt}
\setlist[enumerate]{noitemsep, topsep=1pt, parsep=0pt, partopsep=0pt}

\def\BibTeX{{\rm B\kern-.05em{\sc i\kern-.025em b}\kern-.08em
    T\kern-.1667em\lower.7ex\hbox{E}\kern-.125emX}}

\begin{document}

\title{Forecasting Intraday USD/CAD Exchange Rate with News-Derived Monetary-Policy Signals}

\author{\IEEEauthorblockN{Maya Kodeih\IEEEauthorrefmark{1},
Aliaa Alnaggar\IEEEauthorrefmark{1},
Mucahit Cevik\IEEEauthorrefmark{1}}
\IEEEauthorblockA{\IEEEauthorrefmark{1}Department of Mechanical and Industrial Engineering, 
Toronto Metropolitan University, Toronto, ON, Canada}
\thanks{
Corresponding author(s): M. Kodeih (email: mkodeih@torontomu.ca), A. Alnaggar (email: aliaa.alnaggar@torontomu.ca), M. Cevik (email: mcevik@torontomu.ca).}
}

\maketitle

\begin{abstract}

Monetary-policy announcements and central-bank communications play a central role in foreign exchange markets, yet their qualitative, unstructured form makes their forecasting value difficult to quantify. While prior research has largely focused on sentiment extracted from financial news, comparatively little is known about the relative contribution of different dimensions of monetary-policy communication. Existing studies primarily evaluate whether textual information improves overall forecasting performance but provide limited insight into which communication channels drive such improvements. To address this gap, this paper introduces a statistical attribution methodology that decomposes monetary-policy communication into interpretable channels and quantifies their incremental forecasting contribution under false-discovery-rate control.
Monetary-policy news is transformed into structured communication signals using large language models (LLMs) and temporal feature engineering. These signals are evaluated using rolling-window experiments with tree-based machine-learning models.
The results show that monetary-policy communication contains measurable predictive information. Attribution analysis shows that predictive value is concentrated in a small subset of signals, with communication timing providing the strongest individual feature-level contribution, targeted communication-activity measures also contributing positively, and LLM-derived sentiment providing complementary information at the group level.
The findings indicate that communication-based forecasting value extends beyond sentiment alone and that attribution, rather than aggregate accuracy alone, is central to evaluating news-derived signals.

\end{abstract}


\section{Introduction}
\label{sec:Introduction}

Financial news and central-bank communications play an important role in shaping expectations in foreign exchange (FX) markets. Monetary-policy announcements, policy guidance, and macroeconomic news influence exchange rates by affecting expectations regarding future interest rates, inflation, and economic conditions \cite{blinder2008central,ehrmann2007communication}. Consequently, researchers have increasingly incorporated textual information into FX forecasting systems to complement traditional economic and market-based predictors \cite{tetlock2007giving,bollen2011twitter,loughran2011liability}.
Recent advances in machine learning and financial natural language processing have expanded the methodologies available for financial forecasting. 

Previous work has used tree-based ensembles to capture nonlinear structure in financial time series and large language models (LLMs) to interpret financial text beyond dictionary-based sentiment \cite{ke2017lightgbm, araci2019finbert, wu2023bloomberggpt, yang2023fingpt}. 
Despite these advances, existing studies have primarily evaluated textual information through its effect on aggregate forecasting accuracy \cite{tetlock2007giving,bollen2011twitter,kucuklerli2024sentiment}. While such studies suggest that financial news contains predictive value, comparatively little attention has been devoted to understanding which dimensions of monetary-policy communication drive forecasting improvements. 

Modern forecasting systems often combine sentiment scores with communication activity, policy classifications, news-arrival indicators, temporal persistence measures, and other engineered representations derived from central-bank communications \cite{loughran2011liability,araci2019finbert,wu2023bloomberggpt,yang2023fingpt}. As these feature sets grow more complex, it becomes increasingly difficult to determine whether predictive gains originate from sentiment itself, communication timing, communication frequency, policy stance, or a relatively small subset of engineered communication signals. Although interpretable machine learning has emphasized the importance of feature attribution and model interpretation \cite{lundberg2017shap,molnar2022interpretable,fisher2019model}, systematic empirical evaluation of the relative forecasting value of different dimensions of monetary-policy communication remains limited in FX forecasting.
We address this gap by developing a methodology for attributing the forecasting value of monetary-policy communication in intraday USD/CAD exchange-rate forecasting. Federal Reserve and Bank of Canada news articles are transformed into structured communication signals using an LLM-based sentiment-labeling pipeline and temporal feature engineering. The resulting feature set captures several dimensions of monetary-policy communication, including sentiment, communication activity, communication timing, policy stance, and information persistence. These signals are evaluated using rolling-window forecasting experiments, benchmark comparisons, feature-level and group-level attribution analysis, and false-discovery-rate (FDR)-controlled statistical testing. Rather than asking whether sentiment alone improves forecasting performance, the proposed methodology identifies which dimensions of monetary-policy communication provide measurable and statistically supported forecasting value. Accordingly, the principal methodological contribution is a statistically controlled attribution framework for identifying which dimensions of monetary-policy communication provide incremental forecasting information beyond aggregate predictive performance.

The contributions of this paper are as follows:
\begin{itemize}
\item We present an end-to-end methodology for transforming monetary-policy communications into structured forecasting signals using LLM-based sentiment extraction and temporal feature engineering spanning sentiment, communication timing, communication activity, policy stance, and information persistence.
\item We combine rolling-window forecasting, benchmark comparison, feature-level and group-level attribution analysis, and FDR-controlled statistical testing to quantify the forecasting contribution of individual communication signals.
\item We provide an empirical analysis showing that the forecasting value of monetary-policy communication is concentrated within a small subset of engineered features, with communication timing providing the largest individual feature-level contribution, targeted communication-activity and non-neutral policy signals also contributing positively, and LLM-derived sentiment providing complementary predictive information at the group level.
\end{itemize}

\section{Literature Review}
\label{sec:litReview}

Foreign exchange forecasting remains one of the most challenging problems in finance because exchange rates are influenced by complex interactions among macroeconomic fundamentals, market expectations, and rapidly evolving information flows. Traditional forecasting approaches rely on economic relationships such as purchasing power parity, interest-rate differentials, and monetary fundamentals to explain exchange-rate movements \cite{rossi2013exchange}. Despite decades of research, exchange-rate prediction remains difficult, and many economic models have struggled to consistently outperform naive random-walk benchmarks \cite{cheung2005empirical, rossi2013exchange}.

Recent advances in machine learning have expanded the range of forecasting approaches beyond traditional econometric models. Tree-based ensemble methods, including Random Forests, XGBoost, and LightGBM, have been used to model nonlinear relationships among financial variables and have been applied to exchange-rate forecasting \cite{plakandaras2015forecasting, yildirim2021forecasting}. Recent USD/CAD forecasting studies have further shown that machine-learning models combined with interpretability techniques can extract information from macroeconomic and financial variables such as oil prices, interest rates, equity indices, producer prices, monetary aggregates, unemployment, and industrial production \cite{neghab2025}. However, these studies focus primarily on structured macro-financial variables at daily and weekly frequencies rather than intraday textual information or central-bank communications.

Financial news provides an additional source of information that may influence market expectations before it is fully reflected in prices. Numerous studies have incorporated news and sentiment indicators into forecasting models for equities, commodities, and foreign exchange markets \cite{tetlock2007giving, bollen2011twitter, loughran2011liability}. Central-bank communications are particularly relevant because they convey forward-looking information regarding future monetary-policy actions and can therefore influence exchange-rate expectations before formal policy decisions are implemented \cite{blinder2008central, ehrmann2007communication}. To incorporate textual information into quantitative forecasting systems, researchers commonly transform financial text into sentiment scores, topic representations, or policy-stance classifications. More recently, domain-specific LLMs, including FinBERT, BloombergGPT, and FinGPT, have expanded the ability to extract contextual information from financial communications and have been applied to financial sentiment analysis and market prediction tasks \cite{araci2019finbert, liu2021finbert, wu2023bloomberggpt, yang2023fingpt}.

As forecasting models become increasingly sophisticated, understanding why models perform well has become an important research objective. Interpretable machine-learning techniques such as permutation importance, SHAP, and feature ablation provide complementary approaches for assessing the contribution of individual predictors and distinguishing informative variables from redundant or noisy features \cite{lundberg2017shap, molnar2022interpretable, fisher2019model}. In financial forecasting applications, however, these methods are typically used to explain model behaviour after training rather than to systematically quantify the forecasting value of different categories of engineered communication signals under formal statistical inference.
Consequently, an attribution gap remains. Existing studies generally compare forecasting systems with and without textual information, providing limited insight into whether predictive improvements arise from sentiment itself, communication timing, communication activity, policy stance, temporal persistence, or other engineered representations derived from central-bank communications. 

In contrast to prior USD/CAD work on daily and weekly macro-financial variables \cite{neghab2025}, we study intraday forecasting from news- and LLM-derived central-bank communication signals. Beyond point forecasting, we evaluate directional and probabilistic forecasts and attribute predictive value to individual communication channels through feature- and group-level ablation with FDR control.

\section{Methodology}
\label{sec:Methodology}

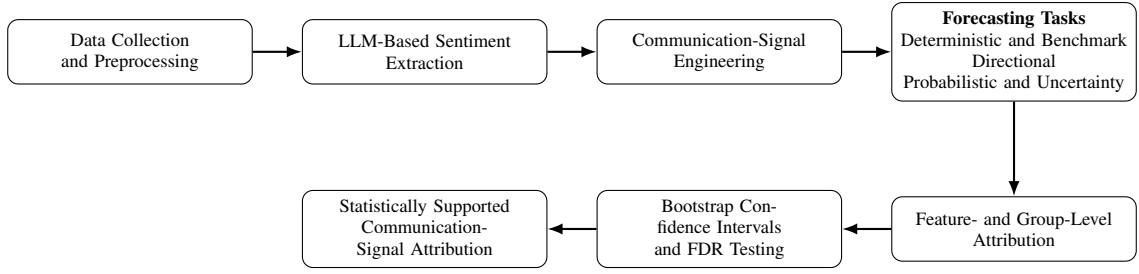
\begin{figure*}[t]
\centering
\begin{tikzpicture}[
box/.style={
rectangle,
draw,
rounded corners,
minimum height=0.85cm,
text width=3.0cm,
align=center,
font=\scriptsize
},
arrow/.style={
-{Latex[length=2mm]},
thick
},
node distance=0.65cm and 0.65cm
]

\node[box] (data)
{Data Collection\\and Preprocessing};

\node[box, right=of data] (llm)
{LLM-Based Sentiment\\Extraction};

\node[box, right=of llm] (features)
{Communication-Signal\\Engineering};

\node[box, right=of features] (forecast)
{\textbf{Forecasting Tasks}\\
Deterministic and Benchmark\\
Directional\\
Probabilistic and Uncertainty};

\draw[arrow] (data) -- (llm);
\draw[arrow] (llm) -- (features);
\draw[arrow] (features) -- (forecast);

\node[box, below=1.25cm of forecast] (attribution)
{Feature- and Group-Level\\Attribution};

\node[box, left=of attribution] (testing)
{Bootstrap Confidence Intervals\\and FDR Testing};

\node[box, left=of testing] (output)
{Statistically Supported\\Communication-Signal Attribution};

\draw[arrow] (forecast.south) -- (attribution.north);

\draw[arrow] (attribution) -- (testing);
\draw[arrow] (testing) -- (output);

\end{tikzpicture}

\caption{Overview of the proposed monetary-policy communication forecasting and attribution workflow.}
\label{fig:framework}
\end{figure*}

We develop an integrated methodology for assessing the forecasting value of monetary-policy communication in the USD/CAD foreign-exchange market. The objective is channel-level attribution rather than aggregate accuracy alone. The methodology consists of seven stages: (1) data collection and preprocessing, (2) LLM-based monetary-policy sentiment extraction, (3) temporal communication-signal engineering, (4) deterministic forecasting and benchmark evaluation, (5) probabilistic forecasting and uncertainty estimation, (6) feature-level and group-level attribution analysis, and (7) statistical significance testing. Figure~\ref{fig:framework} summarizes the overall workflow.

\subsection{Data Collection and Processing}
\label{subsec:data_collection}

Financial news articles were collected from financial news providers and real-time RSS feeds covering macroeconomic developments, monetary-policy decisions, inflation releases, labour-market conditions, and central-bank communications. Because the objective is to evaluate the forecasting value of monetary-policy information, a filtering stage was applied to retain only articles relevant to monetary-policy expectations.

The empirical sample spans 2024-06-13 to 2026-06-12. After filtering and duplicate removal, the final dataset contains 927 monetary-policy-related articles aligned with 17,517 hourly USD/CAD observations. Exchange-rate data were obtained from Yahoo Finance, while news articles were collected through TheNewsAPI with RSS-feed aggregation. The USD/CAD series was aligned to a continuous hourly grid; missing observations, including those during weekend market closures, were forward-filled using the most recent available exchange rate, while any observations provided by Yahoo Finance were retained. Rolling-window evaluation employs an 85/15 train--test split within each window, corresponding to approximately 4,324 training observations and 764 out-of-sample observations, with slightly smaller windows at the beginning of the sample.

Articles were filtered using a monetary-policy vocabulary covering the Federal Reserve, the Bank of Canada, interest rates, inflation, forward guidance, quantitative easing and tightening, and related macroeconomic concepts. Duplicate and near-duplicate articles were removed, and retained articles were timestamped and aggregated to an hourly frequency to align news-derived features with intraday USD/CAD observations.

\subsection{Monetary-Policy Sentiment Extraction}
\label{subsec:sentiment_extraction}

Each filtered article was processed using an LLM to extract monetary-policy sentiment. Sentiment labeling was performed using OpenAI GPT-4o-mini, which classified the implied monetary-policy stance of each article as \emph{hawkish}, \emph{dovish}, \emph{neutral}, or \emph{mixed}. Federal Reserve-related and Bank of Canada-related communications were labeled separately, enabling the construction of institution-specific sentiment measures and policy-divergence signals.

To assess labeling reliability, a random sample of 100 monetary-policy-related articles was manually reviewed and compared with the corresponding LLM classifications. The manual assessment indicated that the generated labels were generally consistent with the monetary-policy stance expressed in the underlying articles within this validation sample. As an additional qualitative validation, periods containing non-neutral central-bank communications were examined using an exploratory event-study analysis. Exchange-rate movements observed around these events were often directionally consistent with the direction implied by the extracted communication signals, providing diagnostic evidence that the labeling pipeline captured economically meaningful information.
The article-level sentiment labels were subsequently aggregated to an hourly frequency and transformed into forecasting features, including sentiment differentials, non-neutral communication indicators, sentiment-persistence measures, and rolling sentiment summaries over multiple lookback windows. 

\subsection{Feature Engineering}
\label{subsec:feature_engineering}

Article-level sentiment observations were transformed into hourly communication signals representing multiple dimensions of monetary-policy information. The engineered feature set includes sentiment-derived variables, communication timing and activity measures, temporal persistence indicators, and conventional market variables, which are summarized in Table~\ref{tab:feature_categories}. 

\begin{table}[!ht]
\centering
\caption{Feature categories used in the forecasting methodology.}
\label{tab:feature_categories}
\begin{tabularx}{\linewidth}{lX}
\toprule
\textbf{Category} & \textbf{Examples} \\
\midrule
Sentiment & Fed sentiment, BoC sentiment, sentiment differential \\
News Activity & News-hour indicators, non-neutral communication events \\
Communication Volume & Fed article counts, BoC article counts, total article volume \\
Temporal & Impulse variables, decay features, rolling sentiment measures \\
Market & USD/CAD lags, technical indicators, volatility measures \\
\bottomrule
\end{tabularx}
\end{table}

Representative engineered variables include rolling communication counts (e.g., \texttt{fed\_articles\_24h}), news-arrival indicators (e.g., \texttt{is\_news\_hour}), sentiment-differential measures, impulse and decay features, and rolling sentiment summaries computed over multiple lookback windows. Additional market variables comprise USD/CAD lags, intraday volatility, the normalized U.S.--Canada interest-rate spread, macroeconomic momentum measures, and WTI crude oil prices.

A central feature used throughout the analysis is the monetary-policy sentiment differential between the Federal Reserve and the Bank of Canada:

\begin{equation}
SD_t = S_t^{Fed} - S_t^{BoC},
\label{eq:sentdiff}
\end{equation}
where $S_t^{Fed}$ and $S_t^{BoC}$ denote the corresponding central-bank sentiment measures at time $t$.
To capture immediate information-arrival effects, sentiment impulse variables were constructed as
\begin{equation}
I_t =
\begin{cases}
SD_t, & \text{if a policy-related article arrives at time } t, \\
0, & \text{otherwise}.
\end{cases}
\label{eq:impulse}
\end{equation}
The persistence of monetary-policy information was modeled using exponentially decaying sentiment variables:
\begin{equation}
D_t = SD_t \exp(-\lambda \Delta t),
\label{eq:decay}
\end{equation}
where $\Delta t$ denotes elapsed time since publication and $\lambda$ controls the decay rate.

Rolling sentiment indicators were constructed using multiple lookback windows:
\begin{equation}
RS_t^{(k)} =
\frac{1}{k}
\sum_{i=0}^{k-1} SD_{t-i},
\label{eq:rolling}
\end{equation}
where $k$ denotes the rolling-window length.

\subsection{Forecasting Models}
\label{subsec:forecasting_models}

Tree-based ensemble models were employed throughout the forecasting framework because of their ability to capture nonlinear relationships, higher-order feature interactions, and complex dependencies without requiring strong distributional assumptions.
The evaluated models were Extra Trees (ETR), XGBoost, and LightGBM.

For deterministic forecasting, separate regression models were trained independently for each forecasting horizon. This direct forecasting strategy avoids recursive error propagation by estimating an individual model for each prediction horizon. For a forecasting horizon of $h$ hours, the models predict the future USD/CAD exchange-rate level, $\hat{y}_{t+h}$, using only information available at time $t$. The same model families were also used for directional classification, where the objective is to estimate the probability of upward or downward USD/CAD movements over each forecasting horizon. 
Probabilistic forecasting extends the deterministic models through quantile regression, with conformal calibration evaluated separately as a post-processing procedure. Details of the probabilistic evaluation are provided in Section~\ref{subsec:probabilistic_forecasting}.

Forecasting performance is evaluated relative to two benchmark strategies. The first is a naive moving-average predictor computed from the three most recent observations within each rolling training window. The second is the no-change random-walk forecast, $\hat{y}_{t+h}=y_t$, which serves as the principal benchmark because of the strong persistence exhibited by exchange-rate levels. These benchmarks contextualize level-forecast accuracy; the paper's main objective remains attribution rather than benchmark dominance.

\subsection{Directional Forecast Evaluation}
\label{subsec:directional_eval}

We also evaluate whether monetary-policy communication helps predict the direction of future USD/CAD movements. 
For a forecast horizon of $h$ hours, the directional target is defined as
\begin{equation}
d_{t+h}=
\begin{cases}
1, & \text{if } y_{t+h}-y_t > 0,\\
0, & \text{otherwise}.
\end{cases}
\label{eq:direction_target}
\end{equation}
where $y_t$ denotes the USD/CAD exchange-rate level at time $t$. Classification models estimate the probability of an upward movement, which is converted into a binary prediction using a decision threshold.

Performance is evaluated using directional accuracy (DA), balanced accuracy, macro F1-score, and the area under the receiver operating characteristic curve (AUC). Because the directional target exhibits class imbalance, results are compared with two benchmark strategies:
\begin{itemize}
\item \textit{Majority-class baseline}: always predicts the most frequently observed directional class.
\item \textit{Persistence baseline}: predicts that the future direction will equal the most recently observed direction.
\end{itemize}
Reporting benchmark-relative performance together with balanced accuracy helps distinguish whether directional improvements reflect predictive signal rather than class imbalance or simple persistence.

\subsection{Probabilistic Forecast Evaluation}
\label{subsec:probabilistic_forecasting}

Deterministic forecasting produces a single estimate of the future USD/CAD exchange-rate level but does not quantify the uncertainty associated with that prediction. To characterize predictive uncertainty, the deterministic forecasting framework was extended to estimate conditional prediction intervals using probabilistic forecasting.

Three probabilistic forecasting approaches were evaluated. LightGBM and XGBoost employ native quantile-regression objectives to estimate conditional quantiles directly, while Extra Trees estimates empirical conditional quantiles from the distribution of predictions across ensemble members. Multiple quantile levels were estimated to construct prediction intervals across all forecasting horizons.

To distinguish probabilistic forecasting quality from interval calibration, the primary comparison is performed using the uncalibrated quantile forecasts. Probabilistic performance is evaluated using pinball loss, weighted interval score (WIS), and empirical prediction-interval coverage, which assess quantile accuracy, overall probabilistic forecast quality, and interval calibration, respectively. Conformal calibration is evaluated separately as a post-processing procedure using held-out calibration residuals and is not included in the feature-specification comparison reported in Table~\ref{tab:probabilistic_step7_comparison}.

\subsection{Feature Attribution Analysis}
\label{subsec:feature_attribution}

To quantify the contribution of individual communication signals, forecasting performance was evaluated using a sequential feature-ablation procedure. Unlike model-specific feature-importance measures, which describe how a fitted model uses available predictors, feature ablation measures the change in out-of-sample forecasting performance after systematically removing information from the forecasting system. This approach estimates the incremental predictive value of each feature under the full forecasting pipeline. Consequently, variables that appear highly important within a fitted model may contribute little incremental forecasting information once correlated communication signals are available, whereas variables with modest model importance may provide unique predictive value. This distinction motivates the cumulative ablation approach adopted in this study.

Beginning with the full feature set, features were removed sequentially according to a predefined ordering. After each removal, the forecasting models were retrained and evaluated using the same rolling-window experimental design. The performance difference between consecutive ablation steps estimates the marginal contribution of the removed feature while allowing downstream feature interactions to adjust following retraining.

To complement the feature-level analysis, related variables were grouped into broader feature groups, including communication timing, communication activity, sentiment, temporal persistence, policy stance, and conventional market variables. Group-level ablation evaluates the collective contribution of each feature group by jointly removing all features within the corresponding category and comparing forecasting performance with the baseline model.

The primary attribution metric is the degradation in out-of-sample forecasting performance relative to the preceding ablation step. Positive degradation indicates that the removed feature or feature group contributed useful predictive information, whereas negligible changes suggest redundancy. Negative degradation indicates that removing the feature improves forecasting performance, implying that the corresponding information is redundant or detrimental within the forecasting system.
Feature-level attribution provides fine-grained estimates of the contribution of individual engineered variables, while group-level attribution characterizes the relative importance of broader feature groups. 

\subsection{Statistical Significance Testing}
\label{subsec:statistical_testing}

We test each ablation effect with bootstrap confidence intervals and Benjamini--Hochberg false-discovery-rate (FDR) correction.
For each feature-level and group-level ablation experiment, rolling-window performance differences were resampled using nonparametric bootstrap procedures to estimate the sampling distribution of the attribution metric. Bootstrap confidence intervals were then computed for each feature and feature group.

Because multiple hypothesis tests were performed simultaneously, statistical significance was evaluated using the Benjamini--Hochberg false-discovery-rate procedure \cite{benjamini1995fdr}. For feature-level attribution, the global FDR adjustment applies the Benjamini–Hochberg procedure across the two-sided p-values of all individual feature-ablation hypotheses, where each feature-level p-value is computed from its pooled stepwise $\Delta R^2$ observations across the evaluated model-horizon combinations. This approach controls the expected proportion of false discoveries while providing greater statistical power than traditional family-wise error-rate corrections.

Features were considered statistically supported only when both conditions were satisfied:
\begin{itemize}
\item the bootstrap confidence interval excluded zero; and
\item the corresponding FDR-adjusted $q$-value satisfied the predefined significance threshold.
\end{itemize}

Applying these criteria makes the reported attribution effects more conservative after accounting for multiple comparisons. Consequently, the analysis identifies communication signals that show consistent predictive contributions across rolling forecasting windows while reducing the likelihood of false-positive discoveries.

\subsection{Experimental Design}
\label{subsec:experimental_design}

Experiments were conducted using a benchmark-aware rolling-window evaluation designed to simulate realistic forecasting conditions. For each evaluation window, models were trained using historical observations and evaluated on subsequent out-of-sample data, so every prediction was generated using only information available at the forecasting time, avoiding look-ahead bias.

The experimental design comprises three complementary prediction tasks. The primary task forecasts future USD/CAD exchange-rate levels using deterministic regression models and forms the basis of the feature-level and group-level attribution analyses. The second task evaluates predictive uncertainty through probabilistic forecasting by estimating conditional prediction intervals. The third task investigates directional predictability using a binary-classification framework. 

Forecast horizons of 1, 3, 8, and 24 hours were evaluated to capture both immediate and persistent communication effects. Deterministic forecasting was assessed using Root Mean Squared Error (RMSE), Mean Absolute Percentage Error (MAPE), and the coefficient of determination ($R^{2}$), probabilistic forecasting using quantile and interval-based scoring metrics, and directional forecasting using standard classification metrics together with majority-class and persistence benchmarks.

Feature-level and group-level attribution analyses are performed using deterministic point forecasts to quantify the incremental contribution of individual communication signals. Probabilistic forecasting is evaluated separately to determine whether the communication hierarchy identified through deterministic attribution remains consistent under uncertainty-aware forecasting.

All experiments were implemented in Python using publicly available machine-learning libraries. Data preprocessing employed \texttt{pandas} and \texttt{NumPy}, while model development used \texttt{scikit-learn}, \texttt{LightGBM}, and \texttt{XGBoost}. LLM sentiment extraction was performed using the OpenAI GPT-4o-mini API.

To support reproducibility, a fixed random seed was used for all experiments. Model hyperparameters remained constant across rolling-window evaluations, with only the training and testing data changing as the evaluation window advanced.

All forecasting experiments followed the same preprocessing pipeline, feature-engineering procedure, rolling-window evaluation protocol, benchmark comparisons, attribution methodology, and statistical testing framework. Performance metrics were computed using identical evaluation procedures across all forecasting horizons and model configurations to allow a fair comparison of deterministic, probabilistic, and directional forecasting results.

\section{Results}
\label{sec:Results}

This section reports deterministic, probabilistic, and directional forecasting results as well as the attribution outcomes. 

\subsection{Forecasting Performance Across Horizons}
\label{subsec:forecasting_performance}

We first use level forecasts to contextualize model fit and support the subsequent attribution analysis. 
Across various forecast horizons, the models achieve high out-of-sample $R^{2}$ (0.844--0.915; Table~\ref{tab:benchmark_comparison}), highest at the 1-hour horizon (LightGBM Direct) and declining at longer horizons (Extra Trees Direct). 
Although these results show high predictive power, they should be interpreted relative to the no-change random-walk benchmark, which has long served as the principal reference model in foreign-exchange forecasting. Table~\ref{tab:benchmark_comparison} compares the best-performing machine-learning model at each horizon with the corresponding random-walk forecast using normalized RMSE, and $R^{2}$.

\begin{table}[!ht]
\centering
\caption{Benchmark comparison for level forecasting.}
\label{tab:benchmark_comparison}
\begin{tabular}{lcccc}
\toprule
\multirow{2}{*}{\textbf{Horizon}} & \multicolumn{2}{c}{\textbf{Best ML Model}} & \multicolumn{2}{c}{\textbf{RW}} \\
\cmidrule(lr){2-3} \cmidrule(lr){4-5}
 & \textbf{$R^2$} & \textbf{NRMSE} & \textbf{$R^2$} & \textbf{NRMSE} \\
\midrule
1h  & 0.9150 & 0.00549 & 0.9988 & 0.00065 \\
3h  & 0.9039 & 0.00584 & 0.9967 & 0.00109 \\
8h  & 0.8886 & 0.00628 & 0.9916 & 0.00173 \\
24h & 0.8442 & 0.00742 & 0.9764 & 0.00289 \\
\bottomrule
\end{tabular}
\end{table}

Consistent with the longstanding foreign-exchange forecasting literature, none of the evaluated machine-learning models outperforms the no-change random walk on level-based forecasting metrics. Across all horizons, NRMSE remains higher than the corresponding random-walk error, and the benchmark achieves higher $R^{2}$ values. 
Benchmark-relative performance nonetheless improves with horizon and the relative performance gap narrows as exchange-rate persistence weakens over longer horizons. While this benchmark result limits claims about level-forecast dominance, it does not preclude attribution within the fitted models.

\begin{figure*}[!ht]
\centering
\includegraphics[width=\textwidth]
{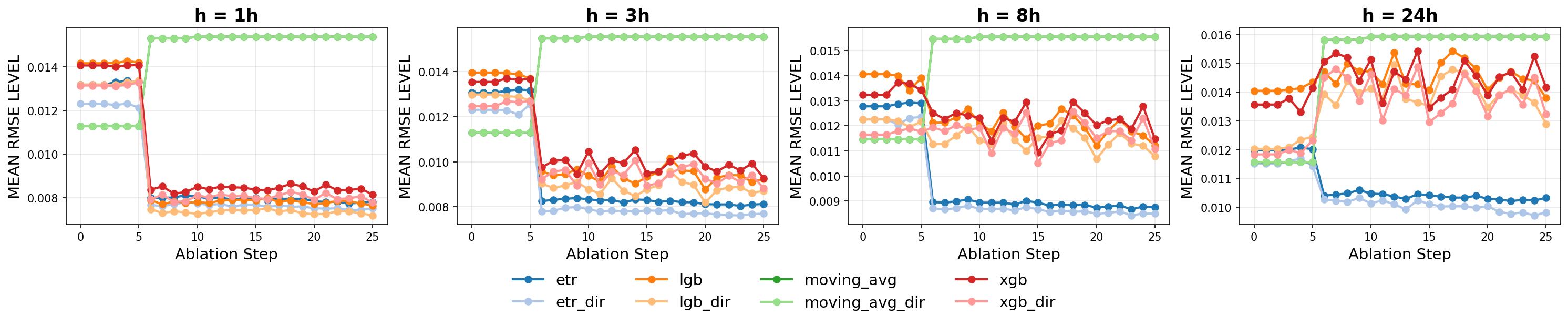}
\caption{Mean RMSE across the cumulative ablation sequence at the 1-, 3-, 8-, and 24-hour horizons. Lower values indicate better forecasting performance. Step~0 is the full feature set; each subsequent step removes one feature.}
\label{fig:ablation_rmse_degradation}
\end{figure*}

Figure~\ref{fig:ablation_rmse_degradation} summarizes the evolution of level-forecasting error throughout the cumulative feature-ablation sequence across the four forecasting horizons. The largest changes occur during the early ablation steps, where several weak or redundant sentiment and directional representations are removed. For the tree-based forecasting models, RMSE generally decreases after these removals, particularly at the 1-hour and 3-hour horizons, indicating that the full feature set contains variables that reduce forecasting efficiency when combined with the more informative communication-timing and communication-activity signals.

At longer horizons, model responses are more heterogeneous. Extra Trees remains comparatively stable throughout the ablation sequence, whereas the boosting models become increasingly sensitive to cumulative feature removal. These results suggest that the contribution of individual communication signals depends not only on the forecasting horizon but also on the interaction among the remaining engineered features.

\subsection{Probabilistic Forecasting Results}
\label{subsec:probabilistic_results}

To evaluate whether the communication hierarchy identified through deterministic attribution also improves uncertainty-aware forecasting, Table~\ref{tab:probabilistic_step7_comparison} compares the full feature set with the Step 7 communication-aware specification using uncalibrated quantile forecasts. The comparison uses LightGBM quantile regression for the Step 7 models and the strongest full-feature probabilistic model at each forecasting horizon. Conformal calibration is evaluated separately and is therefore not reflected in the coverage values reported in Table~\ref{tab:probabilistic_step7_comparison}.

Across all forecasting horizons, the empirical prediction intervals are under-covered relative to their nominal confidence levels. Coverage of the nominal 50\% intervals ranges from 30.85\% to 44.01\%, while coverage of the nominal 90\% intervals ranges from 53.72\% to 63.23\%. These results indicate that the uncalibrated prediction intervals are systematically too narrow, motivating the separate evaluation of conformal calibration as a post-processing procedure.

\begin{table}[!ht]
\centering
\caption{Full-feature versus Step 7 probabilistic forecasting performance using uncalibrated quantile forecasts. Lower pinball loss and WIS indicate better probabilistic performance; Cov90 reports empirical coverage of the nominal 90\% interval before conformal calibration.}
\label{tab:probabilistic_step7_comparison}
\setlength{\tabcolsep}{4.5pt}
\footnotesize
\resizebox{0.99\linewidth}{!}{
\begin{tabular}{lcccccc}
\toprule
\multirow{2}{*}{\textbf{H}} & \multicolumn{2}{c}{\textbf{Pinball Loss}} & \multicolumn{2}{c}{\textbf{WIS}} & \multicolumn{2}{c}{\textbf{Cov90 (\%)}} \\
\cmidrule(lr){2-3} \cmidrule(lr){4-5} \cmidrule(lr){6-7}
 & \textbf{Full} & \textbf{Step 7} & \textbf{Full} & \textbf{Step 7} & \textbf{Full} & \textbf{Step 7} \\
\midrule
1h  & 0.003329 & 0.001479 & 0.002394 & 0.001062 & 53.72 & 68.92 \\
3h  & 0.003731 & 0.001989 & 0.002682 & 0.001429 & 63.23 & 58.47 \\
8h  & 0.004001 & 0.002829 & 0.002877 & 0.002032 & 61.85 & 50.88 \\
24h & 0.004600 & 0.004102 & 0.003308 & 0.002946 & 57.85 & 46.29 \\
\bottomrule
\end{tabular}
}
\end{table}

The Step~7 specification consistently improves the probabilistic scoring rules across all forecasting horizons. Relative to the full feature set, pinball loss decreases by approximately 55.6\% at the 1-hour horizon, 46.7\% at the 3-hour horizon, 29.3\% at the 8-hour horizon, and 10.8\% at the 24-hour horizon. Weighted interval score exhibits a similar pattern, indicating that the feature-removal sequence identified through deterministic attribution also improves probabilistic forecasting performance.
Thus, the ablation hierarchy is informative for probabilistic scoring as well as point forecasting.

The interpretation, however, requires an important qualification. Although Step~7 reduces pinball loss and WIS, it does not uniformly improve empirical interval coverage. Coverage of the nominal 90\% interval increases from 53.72\% to 68.92\% at the 1-hour horizon but decreases at the longer forecasting horizons. Consequently, the Step~7 models produce sharper and better-scoring probabilistic forecasts under proper scoring rules, but they also become less conservative as the forecasting horizon increases.

\begin{figure*}[!ht]
\centering
\begin{minipage}{0.49\textwidth}
\centering
\includegraphics[width=\linewidth]
{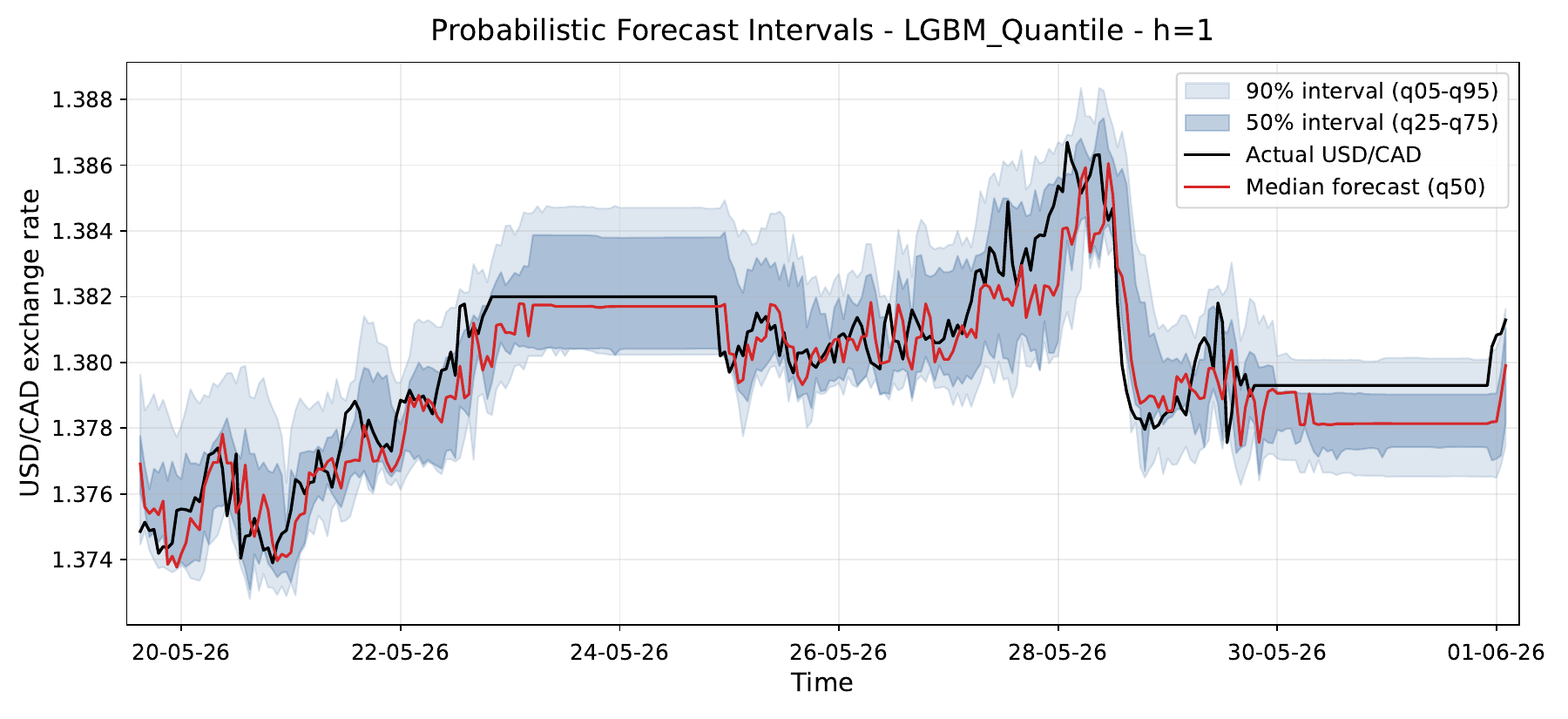}
{\footnotesize (a) 1-hour forecast horizon}
\end{minipage}
\hfill
\begin{minipage}{0.49\textwidth}
\centering
\includegraphics[width=\linewidth]
{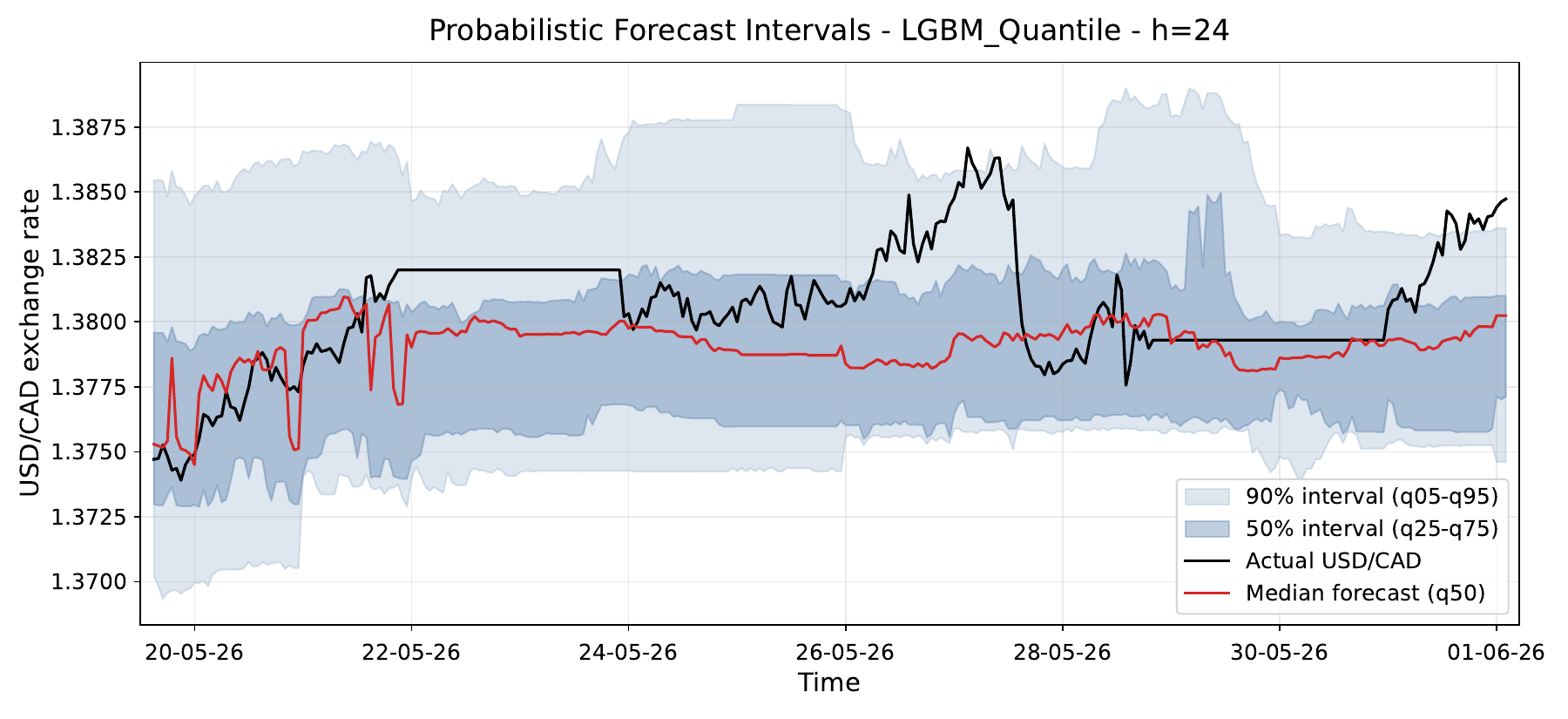}
{\footnotesize (b) 24-hour forecast horizon}
\end{minipage}
\caption{Step~7 LightGBM quantile forecasts at the 1- and 24-hour horizons. Black shows the observed USD/CAD exchange rate, red the median forecast, and shaded regions the 50\% and 90\% prediction intervals. Responsiveness and empirical coverage decline at longer horizons.}
\label{fig:step7_probabilistic_intervals}
\end{figure*}

Figure~\ref{fig:step7_probabilistic_intervals} illustrates these differences across forecast horizons. At the 1-hour horizon, the median forecast closely follows the observed USD/CAD exchange rate, and the prediction intervals capture much of the short-term variation. By the 24-hour horizon, the median forecast responds less effectively to rapid market movements, and observed values fall outside the prediction intervals more frequently. This behaviour is consistent with the reduced empirical coverage reported in Table~\ref{tab:probabilistic_step7_comparison} and reflects the increasing uncertainty associated with longer forecasting horizons.

\subsection{Directional Forecasting Performance}
\label{subsec:directional_performance}

Directional forecasting tests whether communication features are more informative for market direction than for precise price levels.
Table~\ref{tab:directional_performance} summarizes the strongest directional-classification results across the four forecasting horizons. In addition to directional accuracy (DA), the evaluation includes majority-class and persistence benchmarks together with balanced accuracy, macro F1-score, and the area under the receiver operating characteristic curve (AUC). These complementary metrics provide a more balanced assessment because the directional target exhibits varying degrees of class imbalance across forecasting horizons.

\setlength{\tabcolsep}{2pt}
\renewcommand{\arraystretch}{1.03}
\begin{table}[!ht]
\centering
\caption{Directional forecasting performance relative to benchmark strategies.}
\label{tab:directional_performance}
\resizebox{0.99\linewidth}{!}{
\begin{tabular}{lcccccc}
\toprule
\textbf{Horizon} &
\textbf{DA (\%)} &
\textbf{Majority (\%)} &
\textbf{Persistence (\%)} &
\textbf{Balanced Acc.} &
\textbf{Macro F1} &
\textbf{AUC} \\
\midrule
1 hour  & 66.08 & 63.05 & 62.61 & 0.721 & 0.660 & 0.744 \\
3 hours & 65.12 & 61.54 & 61.57 & 0.698 & 0.651 & 0.729 \\
8 hours & 66.55 & 57.80 & 61.51 & 0.705 & 0.658 & 0.707 \\
24 hours & 63.22 & 51.14 & 57.70 & 0.626 & 0.604 & 0.634 \\
\bottomrule
\end{tabular}
}
\end{table}

The directional target is moderately imbalanced at the shorter forecasting horizons. At the 1-hour horizon, the out-of-sample test set contains 939 upward movements and 1,602 downward movements, corresponding to a majority-class benchmark accuracy of 63.05\%. The corresponding class distributions at the 3-, 8-, and 24-hour horizons are 38.46\%/61.54\%, 42.20\%/57.80\%, and 51.14\%/48.86\%, respectively. Consequently, directional accuracy is interpreted relative to both the majority-class and persistence benchmarks.

Across all horizons, the directional models exceed both benchmark strategies. Directional accuracy ranges from 63.22\% to 66.55\%, outperforming the majority-class benchmark by approximately 3.0 percentage points at the 1-hour horizon, 3.6 percentage points at the 3-hour horizon, 8.7 percentage points at the 8-hour horizon, and 12.1 percentage points at the 24-hour horizon. Although absolute directional accuracy changes only modestly across horizons, benchmark-relative improvements become progressively larger, indicating that monetary-policy communication retains predictive information even as exchange-rate persistence weakens.

\begin{figure*}[!ht]
\centering
\includegraphics[width=\textwidth]
{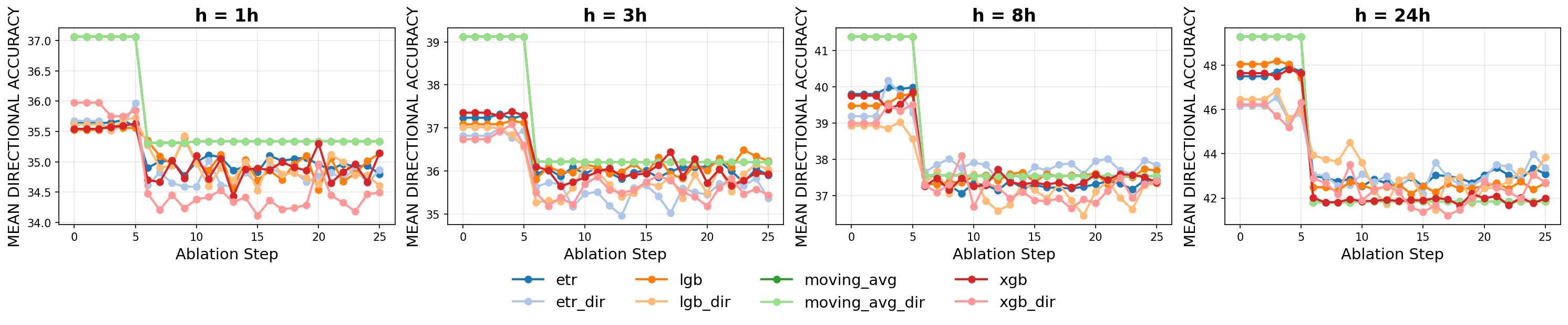}
\caption{Mean directional accuracy across the cumulative ablation sequence at the 1-, 3-, 8-, and 24-hour horizons. Higher values indicate better directional classification. Step~0 denotes the full feature set; each subsequent step removes one additional feature.}
\label{fig:ablation_directional_accuracy}
\end{figure*}

Figure~\ref{fig:ablation_directional_accuracy} shows the evolution of directional accuracy throughout the cumulative feature-ablation sequence. The largest changes occur during the early ablation steps, after which performance remains comparatively stable. These results indicate that the removal of a small number of engineered communication features has a much greater influence on directional performance than subsequent feature removals. The magnitude and direction of these changes vary across forecasting models and horizons, indicating that cumulative ablation effects depend on both the remaining feature set and the forecasting model rather than representing fixed measures of feature importance.

Balanced accuracy ranges from 0.626 to 0.721, above the 0.50 value expected from random directional classification. The strongest balanced-accuracy performance is observed at the 1-hour horizon, while meaningful class discrimination is maintained even at the 24-hour horizon. Similarly, macro F1-scores between 0.604 and 0.660 indicate balanced precision and recall across both directional classes, and AUC values ranging from 0.634 to 0.744 indicate consistent discriminatory ability across alternative decision thresholds.
These complementary metrics indicate that the observed improvements are not attributable to class imbalance alone but reflect directional signal from the engineered monetary-policy communication features. Taken together with the level-forecast benchmark results, these findings suggest that the engineered monetary-policy communication features may be more useful for predicting market direction than for improving precise exchange-rate level forecasts in this setting.

\subsection{Feature-Level Attribution Results}
\label{subsec:feature_attribution_results}

We next examine which engineered communication features drive forecasting performance. Feature-level cumulative ablation quantifies the incremental contribution of each feature by measuring the change in out-of-sample explanatory power after its removal. Changes in $R^{2}$ ($\Delta R^{2}$) serve as the primary attribution metric, while changes in directional accuracy ($\Delta$DA) provide complementary evidence for directional forecasting. Statistical significance is evaluated using the global false-discovery-rate (FDR) correction described in Section~\ref{subsec:statistical_testing}, so that only statistically supported attribution effects are interpreted.

\begin{table*}[!ht]
\centering
\caption{FDR-supported feature-level attribution results. Positive $\Delta R^2$ indicates that feature removal reduces forecasting performance, whereas negative $\Delta R^2$ indicates improved performance after feature removal.}
\label{tab:feature_attribution_results}
\begin{tabular}{lcccc}
\toprule
\textbf{Feature} & \textbf{Mean $\Delta R^2$} & \textbf{Mean $\Delta$DA} & \textbf{95\% CI} & \textbf{Global FDR $q$-value} \\
\midrule

\multicolumn{5}{l}{\textbf{Positive contributors}}\\
\midrule
\texttt{is\_news\_hour}               & +0.012 & +0.04 pp & [+0.006,+0.018] & 0.003 \\
\texttt{fed\_non\_neutral\_last\_24h} & +0.005 & -0.01 pp & [+0.003,+0.008] & 0.003 \\
\texttt{fed\_articles\_24h}           & +0.005 & -0.10 pp & [+0.002,+0.008] & 0.008 \\
\texttt{boc\_news\_last\_24h}         & +0.005 & -0.09 pp & [+0.001,+0.009] & 0.041 \\
\texttt{boc\_articles\_24h}           & +0.003 & +0.08 pp & [+0.001,+0.004] & 0.026 \\

\midrule
\multicolumn{5}{l}{\textbf{Adverse or redundant contributors}}\\
\midrule
\texttt{news\_articles\_24h}          & -0.010 & -0.07 pp & [-0.014,-0.006] & 0.001 \\
\texttt{intraday\_volatility}         & -0.010 & +0.03 pp & [-0.016,-0.004] & 0.008 \\
\texttt{hours\_since\_boc\_news\_168h}& -0.009 & +0.02 pp & [-0.015,-0.004] & 0.008 \\
\texttt{fed\_news\_last\_6h}          & -0.007 & +0.08 pp & [-0.012,-0.004] & 0.008 \\
\texttt{news\_articles\_6h}           & -0.005 & -0.01 pp & [-0.007,-0.002] & 0.008 \\
\texttt{hours\_since\_news\_168h}     & -0.005 & +0.03 pp & [-0.007,-0.002] & 0.007 \\

\bottomrule
\end{tabular}
\end{table*}

Table~\ref{tab:feature_attribution_results} reports the statistically supported feature-level attribution effects. 
The attribution results show that forecasting value is concentrated within a small subset of communication features. The strongest contributor is the indicator identifying hours containing monetary-policy news (\texttt{is\_news\_hour}), whose removal reduces out-of-sample explanatory power by approximately 1.2 percentage points. Although this change appears modest in absolute terms, it should be interpreted relative to the already high baseline forecasting performance of the evaluated models, whose $R^{2}$ values exceed 0.86 across the forecasting horizons.

Beyond communication timing, the remaining statistically significant contributors primarily capture communication activity rather than raw sentiment. Measures of recent Federal Reserve non-neutral communication, Federal Reserve article counts, and Bank of Canada communication activity all contribute positively after global FDR correction. These results indicate that the timing and intensity of central-bank communication provide more reliable forecasting information than broad sentiment measures alone.

The same analysis identifies adverse or redundant variables whose removal improves accuracy.
The adverse contributors exhibit a markedly different pattern. Whereas the strongest positive contributors describe communication timing and communication activity, the adverse features are dominated by aggregate news-volume measures and long-horizon recency variables. In particular, the aggregate article-count variables (\texttt{news\_articles\_24h} and \texttt{news\_articles\_6h}) consistently exhibit negative $\Delta R^{2}$ values, indicating that they become redundant once more targeted Federal Reserve and Bank of Canada communication variables are included. Similarly, long-horizon recency measures such as \texttt{hours\_since\_boc\_news\_168h} and \texttt{hours\_since\_news\_168h} appear to duplicate information already captured by explicit news-arrival and communication-activity indicators.

A clear separation emerges between communication-specific variables, which improve performance, and aggregate news measures, which are often redundant.
Improving communication-based forecasting is thus as much about selecting informative features as about adding new ones.

\subsection{Group-Level Attribution Results}
\label{subsec:group_results}

While the feature-level analysis identifies the individual communication signals responsible for forecasting improvements, group-level attribution evaluates whether broader categories of related features contribute predictive value when considered collectively. This perspective helps characterize how forecasting information is distributed across the principal feature groups, including communication timing, communication activity, and LLM-derived sentiment.

\begin{table}[!ht]
\centering
\caption{Group-level attribution results and representative LLM-derived sentiment features.}
\label{tab:group_attribution_results}
\begin{tabular}{lcc}
\toprule
\textbf{Feature Group / Representative Feature} &
\textbf{Mean $\Delta R^2$} &
\textbf{Mean $\Delta$DA} \\
\midrule

\multicolumn{3}{l}{\textbf{Feature groups}}\\
\midrule
News timing only                  & +0.0005 & -0.01 pp \\
News volume only                  & -0.0018 & -0.01 pp \\
LLM-derived sentiment only        & +0.0020 & +0.02 pp \\
Combined                          & +0.0001 & -0.00 pp \\

\midrule
\multicolumn{3}{l}{\textbf{Representative LLM-derived sentiment features}}\\
\midrule
\texttt{sent\_diff\_impulse}       & +0.001 & -0.05 pp \\
\texttt{sent\_diff\_decay\_24h}    & +0.002 & -0.00 pp \\
\texttt{usd\_cad\_confidence}      & +0.003 & +0.06 pp \\
\texttt{usd\_cad\_direction}       & -0.000 & +0.02 pp \\
\texttt{sent\_diff\_decay\_6h}     & -0.001 & +0.07 pp \\

\bottomrule
\end{tabular}
\end{table}

The grouped attribution results reinforce the feature-level findings while providing a higher-level interpretation of the communication signals. Communication timing exhibits a positive average contribution, consistent with the feature-level result in which \texttt{is\_news\_hour} emerged as the strongest individual contributor. Thus, the timing of policy-relevant information is one of the most informative short-horizon signals.

In contrast, the news-volume group exhibits a negative average contribution. This result is consistent with the adverse feature-level attribution findings, where aggregate article-count variables became redundant once more targeted central-bank communication indicators were included. Measuring the source and timing of policy communication therefore provides more useful information in these experiments than broad measures of overall news volume.

At the group level, LLM-derived sentiment exhibits a positive average contribution, indicating that contextual information extracted from monetary-policy communications provides complementary predictive information beyond communication timing and activity alone.

The lower panel of Table~\ref{tab:group_attribution_results} reports the principal LLM-derived sentiment variables included in the forecasting methodology.
The individual sentiment variables are generally modest in magnitude, with positive contributions for several representative measures. Features such as \texttt{sent\_diff\_decay\_24h} and \texttt{usd\_cad\_confidence} increase average $\Delta R^{2}$, indicating that contextual information extracted from monetary-policy communications complements communication-frequency and timing variables. 

The contribution of the LLM pipeline extends beyond conventional sentiment scoring. It enables the construction of higher-level representations describing policy confidence, directional implications, non-neutral communication, persistence, and sentiment dynamics. Consequently, some activity measures, especially non-neutral communication counts, depend on information extracted during the language-model interpretation stage. Communication timing identifies \emph{when} policy-relevant information becomes available, whereas the LLM-derived features help characterize \emph{what} that information implies for future monetary-policy expectations.
The small contribution of the combined group indicates substantial overlap among the engineered variables.

Taken together, the feature- and group-level attribution results show that predictive information is distributed across multiple communication dimensions. At the feature level, communication timing provides the strongest individual contribution through \texttt{is\_news\_hour}, while targeted central-bank activity measures also contribute positively. At the group level, LLM-derived sentiment provides additional complementary predictive information, whereas broad aggregate news-volume measures are largely redundant.

Finally, cumulative ablation differs sharply from conventional model-based feature importance. Oil price (\texttt{DCOILWTICO}), the rate spread (\texttt{\_Spread\_pct}), and sentiment-decay features rank highest by model importance yet add little under ablation---some even improve performance when removed---whereas \texttt{is\_news\_hour} has low importance but the largest ablation effect.

\section{Conclusion and Discussion}
\label{sec:Conclusion}

This paper evaluated news-derived monetary-policy signals for intraday USD/CAD exchange-rate forecasting. 
The study makes two contributions. First, within the tested framework, predictive value concentrates in a small subset of signals: communication timing provides the strongest individual feature-level contribution, targeted measures of recent central-bank activity also contribute positively, and LLM-derived sentiment provides complementary information at the group level, whereas broad news-volume measures are largely redundant. Second, the study shows that cumulative ablation and conventional feature-importance measures provide fundamentally different perspectives on model behaviour. Feature-importance measures describe how forecasting models utilize available information, whereas cumulative ablation quantifies the incremental forecasting value contributed by each communication signal. This distinction provides a more informative basis for evaluating engineered communication features and for understanding how monetary-policy information contributes to forecasting performance.
The reduced feature set also improves probabilistic scores, though sharpness and interval calibration must be judged separately.

Several limitations should be acknowledged. The empirical evaluation is restricted to the USD/CAD exchange rate and communications from the Federal Reserve and the Bank of Canada, limiting direct generalization to other currencies and monetary-policy regimes. In addition, cumulative ablation is inherently path dependent, so measured marginal contributions may vary with the feature-removal sequence. Finally, the study evaluates statistical forecasting performance rather than economic value and therefore does not consider transaction costs, market frictions, or trading profitability. The LLM-derived communication features are also influenced by prompt design, model selection, and classification uncertainty.

Future research can extend this methodology to additional currency pairs, central banks, and multilingual communication sources to evaluate the stability of the observed communication hierarchy. Alternative attribution approaches, including permutation-based attribution and SHAP-based group explanations, would provide useful comparisons with cumulative ablation. Future work should also incorporate formal benchmark-comparison procedures, such as the Diebold--Mariano and Clark--West tests, together with realistic trading evaluations and more advanced probabilistic calibration techniques.
More broadly, better communication-aware forecasting depends less on adding textual features than on identifying the few that carry unique predictive value.

\small
\bibliographystyle{IEEEtran}
\bibliography{references}

\end{document}